\documentclass[runningheads]{llncs}
\usepackage{csquotes}
\usepackage{graphicx}
\usepackage[ruled,vlined]{algorithm2e}
\usepackage{multirow}
\usepackage{mathtools}
\usepackage{makecell}
\SetKwInput{KwInput}{Input}               
\SetKwInput{KwOutput}{Output} 

\usepackage[usenames,dvipsnames,svgnames,table]{xcolor}

\usepackage{subcaption}
\usepackage{xspace}
\usepackage{amssymb, amsmath}
\makeatletter
\DeclareRobustCommand\onedot{\futurelet\@let@token\@onedot}
\def\@onedot{\ifx\@let@token.\else.\null\fi\xspace}
\def\eg{\emph{e.g}\onedot} 

\def\ie{\emph{i.e}\onedot}

\def\etal{\emph{et al}\onedot}
\makeatother
\DeclareMathOperator{\Loss}{\mathcal{L}}
\DeclareMathOperator*{\R}{\mathbb{R}}

\renewcommand{\vec}[1]{{\mathbf #1}}

\usepackage{hyperref}
\hypersetup{
    colorlinks=true,
    linkcolor=blue,
    filecolor=magenta,      
    urlcolor=cyan,
}

\begin{document}

\title{Duo-SegNet: Adversarial Dual-Views for Semi-Supervised Medical Image Segmentation}

\titlerunning{Duo-SegNet}
\author{Himashi Peiris 
\inst{1}, Zhaolin Chen 
\inst{1,2}, Gary Egan 
\inst{2}, Mehrtash Harandi 
\inst{1}}

\authorrunning{H. Peiris et al.}
%
\institute{Department of Electrical and Computer Systems Engineering, Monash University, Melbourne, Australia. \and Monash Biomedical Imaging (MBI), Monash University, Melbourne, Australia. \\
\email{\{Edirisinghe.Peiris, Zhaolin.Chen, Gary.Egan, Mehrtash.Harandi\}@monash.edu}}
\maketitle 
\begin{abstract}

Segmentation of images is a long-standing challenge in medical AI. This is mainly due to the fact that training a neural network to perform image segmentation requires a significant number of pixel-level annotated data, which is often unavailable. To address this issue, we propose a semi-supervised image segmentation technique based on the concept of multi-view learning. In contrast to the previous art, we introduce an adversarial form of dual-view training and employ a critic to formulate the learning problem in multi-view training as a min-max problem. Thorough quantitative and qualitative evaluations on several datasets, indicate that our proposed method outperforms state-of-the-art medical image segmentation algorithms consistently and comfortably. The code is publicly available at \url{https://github.com/himashi92/Duo-SegNet}. 

\end{abstract}

\section{Introduction}
\label{sec:introduction}
In this paper, we propose a semi-supervised technique based on the concept of multi-view learning~\cite{xu2013survey} to segment medical images. Accurate segmentation of medical images is a key-step in developing Computer-Aided Diagnosis (CAD) and automating various clinical tasks such as image-guided interventions.

The prevailing idea in medical image segmentation is to employ an encoder-decoder structure (\eg, UNet~\cite{ronneberger2015u} and its variants~\cite{zhou2018unet++,oktay2018attention}) and formulate the problem as a dense classification/regression problem, depending on the type of input and the desired output. Most of the existing algorithms are breeze through supervised setting when sizable, annotated datasets are available. However, annotating  large-scale  datasets for image segmentation is challenging and expensive. On one hand, most medical image modalities are hefty in size (\eg, 3D volumes as in MRI and CT) and hence annotation is extremely laborious. On the other hand, annotating medical images requires expert knowledge and cannot be crowd-sourced. Add to this the fact that medical images often contain low contrast slices and ambiguous regions, which in turn makes annotation very difficult. While attaining large annotated datasets is challenging, unlabeled data comes (almost) for free and is abundant. 

Multi-view learning makes use of multiple distinct views of data and benefits from the resulting relationships to achieve accurate models. The principle of consensus~\cite{xu2013survey} states that by minimizing the disagreement on multiple distinct hypotheses, the error of each hypothesis will be minimized.
To be specific, suppose $h^1$ and $h^2$ are two distinct hypothesis defined on a distribution. Under some mild assumptions and as shown in~\cite{dasgupta2002pac}
\[
P(h^1 \neq h^2) \geq \max\big\{ P_{\mathrm{err}}(h^1),P_{\mathrm{err}}(h^2)\big\}\;.
\]
That is,  reducing the disagreement of the two hypotheses minimizes the error of each hypothesis. The consensus principle provides an efficient way with strong theoretical properties to benefit from unlabeled data as shown for example in the celebrated work of Blum and Mitchell~\cite{blum1998combining}. 
Except a handful of studies~\cite{qiao2018deep,peng2020deep}, 
multi-view learning has been mostly studied under the classification regime~\cite{kumar2011co,kiritchenko2001email}. This raises a natural question, \emph{is multi-view learning beneficial when it comes to segmentation?} and \emph{if yes, how it must be formulated?}
Our work takes a step in this direction and provides a way to cultivate unlabeled data for image segmentation. In particular, we propose a dual-view UNet model and equip it with a critic network. Each UNet provides a view of the  data distribution and will be trained by minimizing a supervised loss on the labeled data and a disagreement loss over the unlabeled data. 
The critic is used to facilitate two objectives, \textbf{1.} to ensure that the output of UNets resembles the ground-truth segmentation masks (in the ideal case, the critic is not able to distinguish whether its input is a ground-truth mask or a prediction mask from the UNets), and \textbf{2.} to identify confident parts of a prediction mask to enforce agreement across the views. 
We should stress that unlike classification where (dis)agreement can be formulated readily between predictions, in segmentation, we face a dense prediction problem, meaning a prediction mask includes tens of thousands, if not more, predictions at pixel-level. 
A naive treatment of (dis)agreement loss could lead to inferior results as simply the networks can overfit to agree on the background, which is the dominant part of the prediction mask in many cases. 
Therefore, to train the Duo-SegNet, we formulate the learning as a min-max problem by allowing a critic to stand in as a quantitative subjective referee. Thorough empirical evaluations demonstrate that our method performs well both qualitatively and quantitatively, utilizing small fraction of labeled data.

In short, we have made the following contributions in this work, \textbf{1.} We propose a dual-view learning scheme 
for semi-supervised medical image segmentation. \textbf{2.} We make use of a critic to regularize the training and identify confident parts of a prediction mask towards learning from unlabelled data.

\section{Methodology}
\label{sec:proposed_method}
\begin{figure*}[ht]
\centering
\includegraphics[width=1\linewidth]{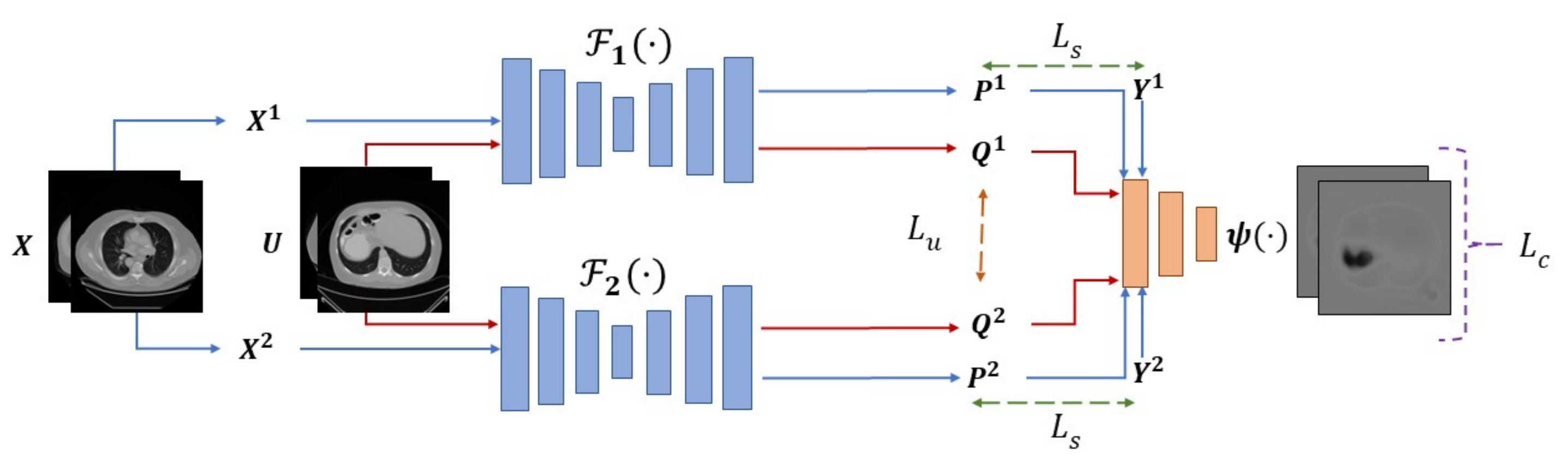} 
\caption{Proposed Adversarial Dual View Network. $\{\mathcal{F}_i(\cdot)\}_{i=1}^2$ and $\psi(\cdot)$ denote Segmentation networks and Critic network. Here, Critic criticizes between prediction masks and the ground truth masks to perform the min-max game.}
\label{fig:architecture}
\end{figure*}
We start this section by providing an overview of the Duo-SegNet (see Fig.~\ref{fig:architecture} for a conceptual illustration). 
Let  $\mathcal{X} = \{(\vec{X}_1,\vec{Y}_1), \cdots , (\vec{X}_m,\vec{Y}_m)\}$ be a labeled set, where each pair $(\vec{X}_i,\vec{Y}_i)$ consists of an image $\vec{X}_i \in \R^{C \times H \times W}$ and its associated ground-truth mask $\vec{Y_i} \in \{0,1\}^{ H \times W}$. Furthermore, let $\mathcal{U} = \{\vec{U}_i\}_{i=1}^n, \vec{U}_i \in \R^{C \times H \times W}$ be a set of $n$ unlabelled images with $n \gg m$. The primary objective is to learn a segmentation model from $\mathcal{D} = \mathcal{X}~\cup~\mathcal{U}$. 

The Duo-SegNet includes two basic modules, a dual view segmentation networks and a critic, shown by $\mathcal{F}_1$, $\mathcal{F}_2$, and $\psi$ respectively in Fig.~\ref{fig:architecture}. Each leg in the dual-view network (\ie, $\mathcal{F}_1$ and $\mathcal{F}_2$) benefits from an encoder-decoder design, and is structured as a UNet~\cite{ronneberger2015u}. In this work, by incorporating a critic, we implicitly enforce segmentation networks to create predictions that are more similar to the desired masks holistically. 
The parameters of the model are the parameters of the dual-view network $\theta_1$, $\theta_2$ and the critic network $\theta_C$. We collectively show the parameters of the dual-view network (\ie, $\theta_1$, $\theta_2$) by $\vec{\Theta}$ to avoid cluttering the equations. To train the model,  we propose optimizing the following min-max problem:
\begin{align}
    \min_{\vec{\Theta}}\max_{\theta_C} \Loss(\vec{\Theta}; \mathcal{D})\;.
    \label{eqn:min_max_overall_loss}    
\end{align}
The familiar min-max problem in~\eqref{eqn:min_max_overall_loss} encourages the dual-view segmentation networks to yield segmentation masks that look like realistic ones by deceiving the critic. 
Similar to standard multi-view training, our approach adapted dual view training to train two segmentation models collaboratively. We propose to train dual view segmentation networks by minimizing a multitask loss function consists of three loss terms:
\begin{align}
    \label{eqn:gen_loss}
    \Loss(\vec{\Theta}; \mathcal{D}) \coloneqq 
    \lambda_s\Loss_{\mathrm{s}}(\vec{\Theta}; \mathcal{X}) 
    + \lambda_u \Loss_{\mathrm{u}}(\vec{\Theta}; \mathcal{U}) + \lambda_c\Loss_{\mathrm{c}}(\vec{\Theta}; \theta_C; \mathcal{D})\;, 
\end{align}
where $\Loss_{\mathrm{s}}$, $\Loss_{\mathrm{u}}$, and $\Loss_{\mathrm{c}}$  denote the supervised, the unsupervised and the critic loss respectively. Furthermore, $\lambda_{\mathrm{s}}, \lambda_{\mathrm{u}}, \lambda_{\mathrm{c}} > 0$ are hyper-parameters of the algorithm, controlling the contribution of each loss term. We note that the supervised and unsupervised loss are only dependent on the dual-view networks while the critic loss is defined based on the parameters of the whole model.  In practice, we find that tuning the hyper-parameters of the network is not difficult at all and the Duo-SegNet works robustly as long as these parameters are defined in a reasonable range. For example, in all our experiments in this paper, we set $\lambda_{\mathrm{s}} = 1.0$, $\lambda_{\mathrm{u}} = 0.3$ and $\lambda_{\mathrm{c}} = 0.2$. 
To better understand Eq.~\eqref{eqn:gen_loss}, we start with the supervised loss. This loss makes use of the labeled data and can be defined as a cross-entropy (\ie, Eq.~\eqref{eqn:ce_seg_loss}) or dice loss (\ie, Eq.~\eqref{eqn:dice_loss}) or even a linear combination of both, which is a common practice in segmentation~\cite{ronneberger2015u}. \begin{align}
    \label{eqn:ce_seg_loss}
    \Loss_{\mathrm{ce}}(\vec{\theta}_i;\mathcal{X}) &= \mathbb{E}_{(\vec{X}, \vec{Y}) \sim \mathcal{X}} \bigg[ 
     \Big\langle \vec{Y}, \log \big(\mathcal{F}_i(\vec{X}) \big) \Big\rangle \bigg]\\
    \label{eqn:dice_loss}
    \hspace{-0.4ex}
    \Loss_{\mathrm{dice}}(\theta_i;\mathcal{X}) \hspace{-0.1ex} &= 1 - \mathbb{E}_{(\vec{X}, \vec{Y}) \sim \mathcal{X}} \Bigg[ \frac{ 2 
     \big \langle \vec{Y} \hspace{-0.1ex}~,~ \hspace{-0.1ex} \mathcal{F}_i(\vec{X}) 
     \hspace{-0.2ex} \big \rangle}
     {\big\|\vec{Y}\big\|_1 + \big\|\mathcal{F}_i(\vec{X})\big\|_1} \Bigg],
\end{align}
where we use $\langle \vec{A},\vec{B}\rangle = \sum_{i,j} \vec{A}[i,j]\vec{B}[i,j]$ 
and $\| \vec{A}\|_1 = \sum_{i,j} |\vec{A}[i,j]|$.

We define the unsupervised loss as a means to realize the principle of consensus as;
\begin{align}
    \label{eqn:unsupervised_loss}
    \Loss_{\mathrm{u}}(\vec{\Theta}; \mathcal{U}) \coloneqq 
    \mathbb{E}_{\vec{U} \sim \mathcal{U}} \bigg[ 
     \Big\langle \mathcal{F}_1(\vec{U}), \log \big(\mathcal{F}_2(\vec{U}) \big) \Big\rangle +
     \Big\langle \mathcal{F}_2(\vec{U}), \log \big(\mathcal{F}_1(\vec{U}) \big) \Big\rangle
     \bigg]\;.
\end{align}
We identify Eq.\eqref{eqn:unsupervised_loss} as a symmetric form of the cross-entropy loss.
In essence, the unsupervised loss $\Loss_{\mathrm{u}}$ acts as an agreement loss during dual-view training. The idea is that the two segmentation networks should generate similar segmentation masks for unlabeled data. In the proposed method, diversity among two views are retrieved based on segmentation model predictions. Unlike in~\cite{peng2020deep,qiao2018deep} where adversarial examples for a model are used to teach other model in the ensemble, we make use of a critic to approach inter diversity among segmentation models for both labeled and unlabeled distributions. When using the labeled data, two segmentation networks are supervised by both the supervised loss with the ground truth and the adversarial loss. We denote the functionality of the critic by $\Psi:[0,1]^{H \times W} \to [0,1]^{H \times W}$ and define the normalized loss of critic for labeled prediction distribution as:
\begin{align}
    \Loss_{{adv1}}(\vec{\Theta};\theta_C; \mathcal{X}) &\coloneqq \mathbb{E}_{(\vec{X}, \vec{Y}) \sim \mathcal{X}} \Bigg[- \sum_{a \in H} \sum_{b \in W} \bigg\{(1-\eta) 
    \log\Big(\psi(\vec{Y})[a,b]\Big)
     \notag \\&+ \eta\log\Big(1 - \psi(\mathcal{F}_i(\vec{X}))[a,b]\Big) 
    \bigg\}\Bigg]\;,
    \label{eqn:loss_critic}
\end{align}
where $\eta = 0$ if the sample is generated by the segmentation network, and $\eta = 1$ if the sample is drawn from the ground truth labels. For unlabeled data, it is obvious that we cannot apply any supervised loss since there is no ground truth annotation available. However, adversarial loss can be applied as it only requires the knowledge about whether the mask is from the ground-truth labels or generated by the segmentation networks. The adversarial loss for the distribution of unlabeled predictions is defined as:
\begin{align}
    \Loss_{{adv2}}(\vec{\Theta};\theta_C; \mathcal{U}) &\coloneqq \mathbb{E}_{\vec{U} \sim \mathcal{U}} \Bigg[-\sum_{a \in H} \sum_{b \in W} \bigg\{ 
    \log\Big(1 - \psi(\mathcal{F}_i(\vec{U}))[a,b]\Big) 
    \bigg\}\Bigg]\;,
    \label{eqn:loss_u_critic}
\end{align}

\begin{remark} 
The basis of having a critic is that a well trained critic can produce pixel-wise uncertainty map/confidence map which imposes a higher-order consistency measure of prediction and ground truth. So, it infers the pixels where the prediction masks are close enough to the ground truth distribution.
\end{remark}

\noindent Therefore, the critic loss is defined as:
 \begin{equation}
     \label{eqn:total_adv_loss}
     \Loss_{\mathrm{c}}(\vec{\Theta}; \theta_C; \mathcal{D}) = \Loss_{\mathrm{adv1}}(\vec{\Theta}; \theta_C; \mathcal{X}) + \Loss_{\mathrm{adv2}}(\vec{\Theta}; \theta_C; \mathcal{U})\;.
 \end{equation}
With this aggregated adversarial loss, we co-train two segmentation networks to fool the critic by maximizing the confidence of the predicted segmentation being generated from the ground truth distribution. To train the critic, we use labeled prediction distribution along with given ground truth segmentation masks and calculate the normalized loss as in Eq.(\ref{eqn:loss_critic}). The overall min-max optimization process is summarized in Algorithm~\ref{alg:duosegnet}.
\begin{algorithm}[ht]
\fontsize{6.5}{7.5}
\SetAlgoLined
\KwInput{Define Segmentation networks $\{\mathcal{F}_i(\cdot)\}_{i=1}^2$, critic $\psi(\cdot)$, batch size $\mathcal{B}$, maximum epoch $E_{max}$, number of steps $k_{s}$ and $k_{c}$ for segmentation networks and critic, Labeled images $\mathcal{X} = \{(X_1,Y_1), ... , (X_m,Y_m)\}$, Unlabeled images $\mathcal{U} = \{U_1, ... , U_n\}$ and two labeled sets $\mathcal{X}^{1}; \mathcal{X}^{2} \subset \mathcal{X}$ ;}
\KwOutput{Network Parameters $\{\theta_i\}_{i=1}^2$ and $\theta_C$;}
 Initialize Network Parameters $\{\theta_i\}_{i=1}^2$ and $\theta_C$\;
 \For{epoch = 1, $\cdots$ , $E_{max}$}{
     \For {batch = 1, $\cdots$ , $\mathcal{B}$}{
         \For {$k_{s}$ steps}{
         Generate predictions for labeled data $\mathcal{F}_1(x)$ for all $X_i \in \mathcal{X}^1$, $\mathcal{F}_2(x)$ for all $X_i \in \mathcal{X}^2$ and for unlabeled data $\mathcal{F}_1(x)$ and $\mathcal{F}_2(x)$ for all $U_i \in \mathcal{U}$;
         
         Generate confidence maps for all predictions using $\psi(\cdot)$;
         
         Let $\Loss = \Loss_{s} + \Loss_u + \Loss_c$, as defined in Equations. (\ref{eqn:gen_loss}) - (\ref{eqn:total_adv_loss});
         
         Update $\{\theta_i\}_{i=1}^2$ by descending its stochastic gradient on $\Loss$;
         }
         
         \For {$k_{c}$ steps}{
         Generate confidence maps for all labeled predictions and ground truth masks using $\psi(\cdot)$;
         
         Let $\Loss_c = \Loss_{adv1}$, as defined in \ref{eqn:loss_critic};
         
         Update $\theta_C$ by ascending its stochastic gradient on $\Loss_c$;
         }
     }
} 
 \caption{Duo-SegNet (\emph{training})}
 \label{alg:duosegnet}
\end{algorithm}
\section{Related Work}
\label{sec:related_work}
We begin by briefly discussing alternative approaches to semi-supervised learning, with a focus on the most relevant ones to our approach, namely pseudo labelling~\cite{lee2013pseudo}, mean teacher model~\cite{tarvainen2017mean}, Virtual Adversarial Training (VAT)~\cite{miyato2018virtual}, recently published deep co-training~\cite{peng2020deep}. Our goal here is to discuss the resemblance and differences between our proposed approach and some of the methods that have been adopted for semi-supervised image segmentation. 

In Pseudo Labelling, as the name implies, the model uses the predicted labels of the unlabeled data and treat it as ground-truth labels for training. The drawback here is, sometimes incorrect pseudo labels may diminish the generalization performance and weaken the training of deep neural networks. In contrast, Duo-SegNet by incorporating a critic, increases the tolerance of these incorrect pseudo labels which stabilizes the generalization performance.
Similar to Duo-SegNet, the mean teacher approach benefits from two neural networks, namely the teacher and the student networks. While the student model is trained in a stochastic manner, the parameters of the teacher model are updated slowly by a form of moving averaging of the student's parameters. This, in turn, results in better robustness to prediction error  as one could hope that averaging attenuates the effect of noisy gradient (as a result of incorrect pseudo labels). Unlike mean teacher model, Duo-SegNet simultaneously train both networks and models can learn from one another during training.
VAT can be understood as an effective regularization method which optimizes generalization power of a model for unlabeled examples. This is achieved by generating  adversarial perturbations to the input of the model, followed by making the model robust to the adversarial perturbations. In contrast to VAT, Duo-SegNet makes use of a critic to judge if predictions are from the same or different distribution compared to labeled examples. With this, segmentation networks are encouraged to generate similar predictive distribution for both labeled and unlabeled data. 
Similar to our work, in co-training two models are alternately trained on  distinct views, while learning from each other is encouraged. Recently, Peng \etal introduced a deep co-training method for semi-supervised image segmentation task~\cite{peng2020deep} based on the approach presented by Qiao \textit{et al.} for image recognition task in~\cite{qiao2018deep}. In this approach diversity among views are achieved via adversarial examples following VAT~\cite{miyato2018virtual}. 
We note that adversarial examples, from a theoretical point of view, cannot guarantee diversity, especially when unlabeled data is considered. That is, a wrong prediction can intensify even more once adversarial examples are constructed from it. 
In contrast, inspired by Generative Adversarial Networks(GANs)~\cite{goodfellow2014generative} and GAN based medical imaging applications including medical image segmentation~\cite{mahmood2019deep}, reconstruction~\cite{quan2017compressed} and domain adaptation~\cite{zhang2018task}, our proposed method graciously encloses the min-max formulation in dual-view learning for segmenting medical images where high-confidence predictions for unlabeled data are leveraged, which is simple and effective.

\section{Experiments}
\label{sec:experiments}
\subsubsection{Implementation Details:}
The proposed model is developed in PyTorch~\cite{pytorch}. Training was done from scratch without using any pre-trained model weights. For training of segmentation network and critic, we use SGD optimizer(LR=1e-02) and RMSProp optimizer(LR=5e-05), respectively. We divide the original dataset into training (80\%) and test set (20\%). Experiments were conducted for 5\%, 20\% and 50\% of labeled training sets.

\subsubsection{Datasets:} We use three medical image datasets for model evaluation covering three medical image modalities : 670 Fluorescence Microscopy (FM) images from Nuclei~\cite{nuclei}, 20 MRI volumes from Heart~\cite{msd} and 41 CT volumes from Spleen~\cite{msd}. For our experiments, 2D images are obtained by slicing the high-resolution MRI and CT volumes for Heart (2271 slices) and Spleen (3650 slices) datasets. Each slice is then resized to a resolution of $256\times256$. 

\subsubsection{Competing Methods and Evaluation Metrics:}
We compare our proposed method with Mean Teacher~\cite{tarvainen2017mean}, Pseudo Labelling~\cite{lee2013pseudo}, VAT~\cite{miyato2018virtual}, Deep Co-training~\cite{peng2020deep} and fully supervised U-Net~\cite{ronneberger2015u}. For all baselines, we follow the same configurations as for our method. All approaches are evaluated using : 1) Dice S$\o$rensen coefficient (DSC) and 2) Mean Absolute Error (MAE).

\subsubsection{Performance Comparison:}
\begin{figure*}[!htb]
\scriptsize
\tabcolsep=0.04cm
\centering
  \begin{tabular}{{c@{ } c@{ } c@{ } c@{ } c@{ }}}
    {\includegraphics[width=0.112\linewidth ]{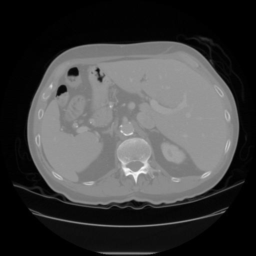}}&
    {\includegraphics[width=0.112\linewidth]{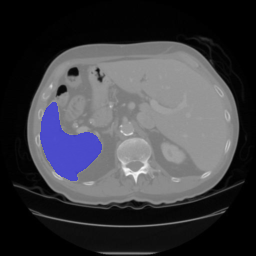}}&
    {\includegraphics[width=0.112\linewidth]{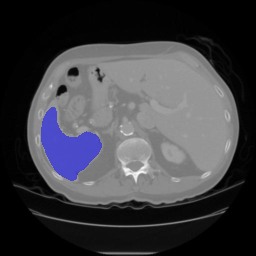}}&
    {\includegraphics[width=0.112\linewidth]{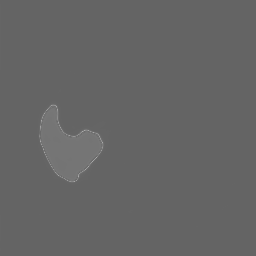}}\\
    \makecell{Input} &
     \makecell{GT} &\makecell{Prediction} &\makecell{Confidence \\ Map} \\
  \end{tabular}
    \caption{Visual Analysis of Confidence Map generated by the Critic during training}
    \label{fig:confidence_maps}
\end{figure*}

\begin{figure*}[h]
\tabcolsep=0.05cm
\scriptsize
\centering
\begin{tabular}{l *{8}{c}}
\rotatebox{90}{Nuclei} & \includegraphics[width=0.112\linewidth]{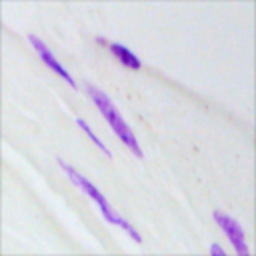} &
\includegraphics[width=0.112\linewidth]{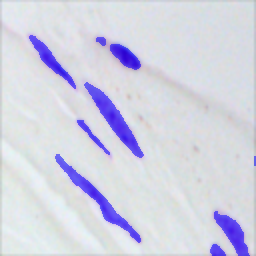}& 
\includegraphics[width=0.112\linewidth]{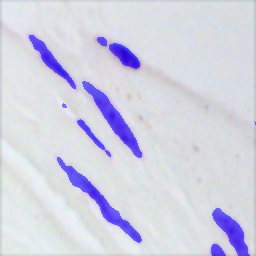}& 
\includegraphics[width=0.112\linewidth]{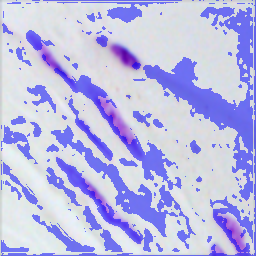}& 
\includegraphics[width=0.112\linewidth]{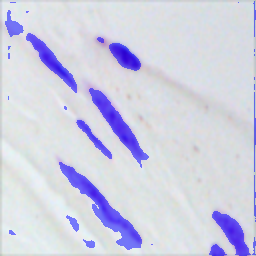}&
\includegraphics[width=0.112\linewidth]{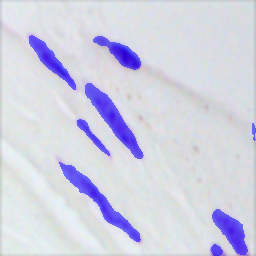}& 
\includegraphics[width=0.112\linewidth]{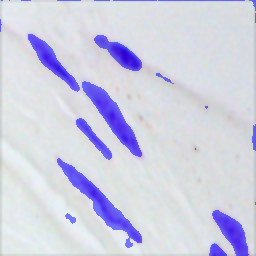}
&
\includegraphics[width=0.112\linewidth]{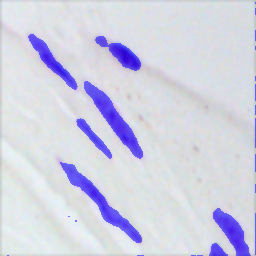}\\
\rotatebox{90}{Heart} & \includegraphics[width=0.112\linewidth, trim={1.5cm 1.5cm 1.5cm 1.5cm},clip]{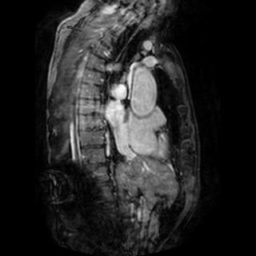} &
\includegraphics[width=0.112\linewidth, trim={1.5cm 1.5cm 1.5cm 1.5cm},clip]{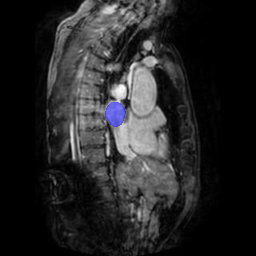}& 
\includegraphics[width=0.112\linewidth, trim={1.5cm 1.5cm 1.5cm 1.5cm},clip]{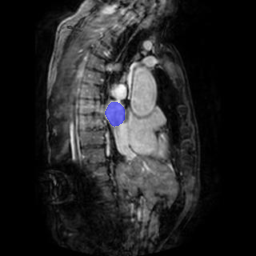}& 
\includegraphics[width=0.112\linewidth, trim={1.5cm 1.5cm 1.5cm 1.5cm},clip]{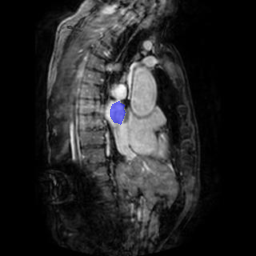}& 
\includegraphics[width=0.112\linewidth, trim={1.5cm 1.5cm 1.5cm 1.5cm},clip]{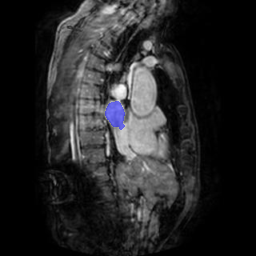}&
\includegraphics[width=0.112\linewidth, trim={1.5cm 1.5cm 1.5cm 1.5cm},clip]{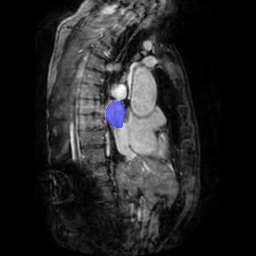}& 
\includegraphics[width=0.112\linewidth, trim={1.5cm 1.5cm 1.5cm 1.5cm},clip]{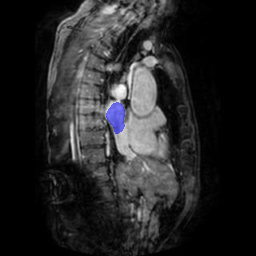}
&
\includegraphics[width=0.112\linewidth, trim={1.5cm 1.5cm 1.5cm 1.5cm},clip]{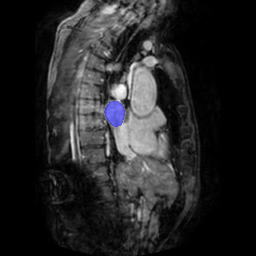}\\

\rotatebox{90}{Spleen} & \includegraphics[width=0.112\linewidth, trim={1cm 1cm 1cm 1cm},clip]{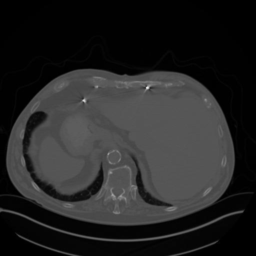} &
\includegraphics[width=0.112\linewidth, trim={1cm 1cm 1cm 1cm},clip]{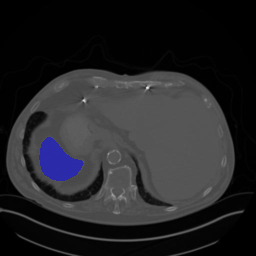}& 
\includegraphics[width=0.112\linewidth, trim={1cm 1cm 1cm 1cm},clip]{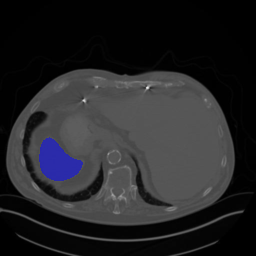}& 
\includegraphics[width=0.112\linewidth, trim={1cm 1cm 1cm 1cm},clip]{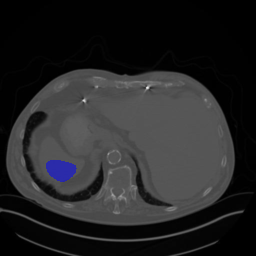}& 
\includegraphics[width=0.112\linewidth, trim={1cm 1cm 1cm 1cm},clip]{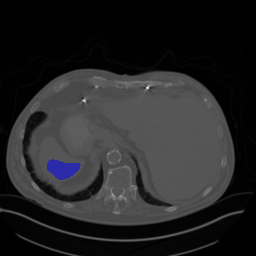}&
\includegraphics[width=0.112\linewidth, trim={1cm 1cm 1cm 1cm},clip]{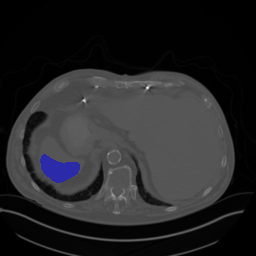}& 
\includegraphics[width=0.112\linewidth, trim={1cm 1cm 1cm 1cm},clip]{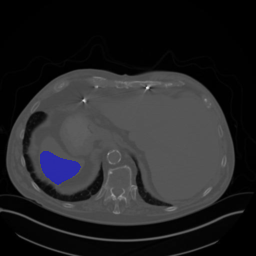}
&
\includegraphics[width=0.112\linewidth, trim={1cm 1cm 1cm 1cm},clip]{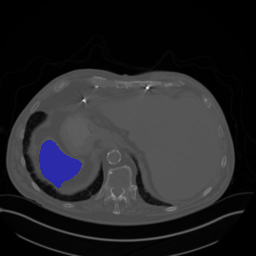}\\
& Input & GT & \makecell{Fully \\ Supervised} & \makecell{Mean \\ Teacher} & \makecell{Pseudo \\ Labelling} & VAT & \makecell{Deep \\ Co-training}& \makecell{Duo-SegNet \\ (Ours)}\\
\end{tabular}
\caption{\textbf{Visual comparison of our method with state-of-the-art models.} Segmentation results are shown for 5\% of labeled training data.}
\label{fig:qualitativeresults}
\end{figure*}

The qualitative results for the proposed and competing methods are shown in Fig.~\ref{fig:qualitativeresults}. The quantitative results comparison of the proposed method to the four state-of-the-art methods are shown in Table~\ref{quantitative}. The results reveal that the proposed method comfortably outperform other studied methods for smaller fractions of annotated data (\eg Spleen 5\%). 
The gap between the Duo-SegNet and other competitors decreases on the Nuclei dataset, when the amount of labeled data increases. That said, we can still observe a significant improvement on the Heart and Spleen dataset. The proposed network can produce both accurate prediction masks and confidence maps representing which regions of the prediction distribution are close to the ground truth label distribution. This is useful when training unlabeled data. Fig.~\ref{fig:confidence_maps} shows the visual analysis of confidence maps. 

\begin{table*}[!htb]
\footnotesize
\centering
\caption{Comparison with state-of-the-art methods.}
\begin{tabular}{l l c c c | c c c}
\hline
Dataset & Method & \multicolumn{3}{c |}{DSC} & \multicolumn{3}{c}{MAE}  \\
\hline
\hline
\multirow{6}{*}{Nuclei} & Fully Supervised & \multicolumn{3}{c |}{\textbf{91.36}}  & \multicolumn{3}{c}{\textbf{2.25}} \\ \cline{3-8}
& & $l_a=5\%$ & $l_a=20\%$ & $l_a=50\%$ & $l_a=5\%$ & $l_a=20\%$ & $l_a=50\%$\\ \cline{3-8}
& Mean Teacher & 83.78 & 84.92 & 87.99 & 4.78 & 4.30& 3.36 \\
& Pseudo Labeling & 60.90 & 72.46 & 85.91 & 8.40 & 6.37 &	3.84 \\ 
& VAT &  85.24 & 86.43 & 88.45 & 4.09 & 3.77 & 3.26\\ 
& Deep Co-training & 85.83 & 87.15 & 89.20 & 4.08 & 3.80 & 3.08 \\
& Duo-SegNet & \textbf{87.14} & \textbf{87.83} & \textbf{89.28} & \textbf{3.57} & \textbf{3.43}  & \textbf{3.03} \\
\hline

\multirow{6}{*}{Heart} & Fully Supervised & \multicolumn{3}{c|}{\textbf{97.17}} & \multicolumn{3}{c}{\textbf{0.02}} \\ \cline{3-8}
& & $l_a=5\%$ & $l_a=20\%$ & $l_a=50\%$ & $l_a=5\%$ & $l_a=20\%$ & $l_a=50\%$\\ \cline{3-8}
& Mean Teacher & 71.00 & 87.59 & 93.43& 0.22 & 0.09 & 0.05 \\ 
& Pseudo Labeling& 65.92 & 79.86 & 80.75 &0.20 & 0.13 & 0.13\\ 
& VAT & 85.33 & 91.60 & 94.83 & 0.11 & 0.06 & 0.04 \\ 
& Deep Co-training & 85.96 & 91.54  & 94.55 & \textbf{0.10} & 0.06 & 0.04 \\
& Duo-SegNet & \textbf{86.79} & \textbf{93.21} & \textbf{95.56} & \textbf{0.10} & \textbf{0.05} & \textbf{0.03}\\
\hline

\multirow{6}{*}{Spleen} & Fully Supervised & \multicolumn{3}{c|}{\textbf{97.89}} & \multicolumn{3}{c}{\textbf{0.02}}  \\ \cline{3-8}
& & $l_a=5\%$ & $l_a=20\%$ & $l_a=50\%$ & $l_a=5\%$ & $l_a=20\%$ & $l_a=50\%$\\ \cline{3-8} 
& Mean Teacher & 75.44 & 90.76 & 92.98 & 0.20 & 0.08 & 0.06 \\
& Pseudo Labeling & 67.70 & 68.81 & 84.81 & 0.24 & 0.21 & 0.12\\ 
& VAT & 78.31 & 91.37 & 94.34 & 0.19 & 0.07 & 0.05\\ 
& Deep Co-training & 79.16 & 89.65 & 94.90 & 0.16 & 0.09 & 0.05 \\
& Duo-SegNet & $\textbf{88.02}$ & $\textbf{92.19}$ & $\textbf{96.03}$ & \textbf{0.10} & \textbf{0.07} & \textbf{0.03} \\
\hline
\end{tabular}
\label{quantitative}
\end{table*}
\subsubsection{Ablation Study:}
We also perform ablation studies to show the effectiveness of adding a critic in dual-view learning in semi-supervised setting and the importance of dual view network structure. In our algorithm, we benefit from unlabeled data via (1) criss-cross exchange of confident regions, (2) improving the critic which in essence minimizes an upper-bound of error. To justify this, we conducted an additional experiment without unlabeled data. It can be seen that there is an impact in the performance of segmentation model for varying values for $\lambda_u$. For our experiments we choose $\lambda_u$ in the range of $0.3$ to $0.4$. All experiments in Table~\ref{ablation} are conducted for spleen dataset with 5\% of annotated data.
\begin{table}[!h]
    \caption{Ablation Study}
    \begin{subtable}{.5\linewidth}
    \footnotesize
      \centering
        \caption{Network Structure Analysis.}
        \begin{tabular}{| l | c | c | }
        \hline
        Experiment & ~DSC~ & ~MAE~ \\
        \hline
        Duo-SegNet & 88.02 & 0.10 \\
        w/o Critic~ & 77.69 & 0.19 \\
        w/o Unlabeled Data & 76.67 & 0.17 \\
        One Segmentation Network~ & 82.44 & 0.16 \\
        \hline
        \end{tabular}
    \end{subtable}%
    \begin{subtable}{.5\linewidth}
    \footnotesize
      \centering
        \caption{Hyper-parameter Analysis for $\lambda_u$.}
        \begin{tabular}{| c | c | c | c | c | c |}
        \hline
        ~$\lambda_u$~ & 0.1 & 0.2 & 0.3 & 0.4 & 0.5 \\
        \hline
        ~DSC~ & 83.58 & 85.62 & 88.02 & 87.14 & 78.89 \\
        \hline
        ~MAE~ & 0.15 & 0.12 & 0.10 & 0.11 & 0.20 \\
        \hline
        \end{tabular}
    \end{subtable} 
\label{ablation}  
\end{table}

\section{Conclusion}
\label{sec:conclusion}
We proposed an adversarial dual-view learning approach for semi-supervised medical image segmentation and demonstrated its effectiveness on publicly available three medical datasets. Our extensive experiments showed that employing a min-max paradigm into multi-view learning scheme sharpens boundaries between different regions in prediction masks and yield a performance close to full-supervision with limited annotations. The dual view training can still be improved by self-tuning mechanisms, which will be considered in our future works.

\bibliographystyle{splncs04}
\bibliography{references}

\end{document}